\documentclass{article}
\usepackage{nips00e,times}
\usepackage{epsfig}
\newcommand{\ignore}[1]{}

\newcommand{\mysection}[1]{\vspace*{-2.2ex}\section{#1}\vspace*{-1.2ex}}
\newcommand{\mysubsection}[1]{\vspace*{-1.2ex}\subsection{#1}\vspace*{-.85ex}}
\newcommand{\mysubsubsection}[1]{\vspace*{-1.2ex}\subsubsection*{#1}\vspace*{-.85ex}}

\title{
The Use of Classifiers in Sequential Inference}

\author{Vasin Punyakanok \hspace*{1.0in} Dan Roth \\
Department of Computer Science \\
University of Illinois at Urbana-Champaign \\
Urbana, IL 61801 \\
{\it punyakan@cs.uiuc.edu \hspace*{0.25in} danr@cs.uiuc.edu} }

\begin{document}
\maketitle
\vspace{-0.1 in}
\begin{abstract}
We study the problem of combining the outcomes of several different
classifiers in a way that provides a coherent inference that satisfies
some constraints. In particular, we develop two general approaches for
an important subproblem - identifying phrase structure.
The first is a Markovian approach that extends standard HMMs to allow the
use of a rich observation structure and of general classifiers to
model state-observation dependencies.
The second is an extension of constraint satisfaction formalisms.  We
develop efficient combination algorithms under both models and study
them experimentally in the context of shallow parsing.
\end{abstract}

%------------------------------------------------------------
\mysection{Introduction} \label{sec:intro}

In many situations it is necessary to make decisions that depend
on the outcomes of several different classifiers in a way that
provides a coherent inference that satisfies some constraints -
the sequential nature of the data or other domain specific
constraints.
Consider, for example, the problem of {\em chunking} natural language
sentences where the goal is to identify several kinds of phrases (e.g.
noun phrases, verb phrases) in sentences.  A task of this sort
involves multiple predictions that interact in some way.  For example,
one way to address the problem is to utilize two classifiers for each
phrase type, one of which recognizes the beginning of the phrase, and
the other its end. Clearly, there are constraints over the
predictions; for instance, phrases cannot overlap and there are
probabilistic constraints over the order of phrases and their lengths.
The above mentioned problem is an instance of a general class of
problems -- identifying the phrase structure in sequential data.
This paper develops two general approaches for this class of
problems by utilizing general classifiers and performing
inferences with their outcomes.  Our formalisms directly applies
to natural language problems such as shallow
parsing~\cite{Church88,RamshawMa95,CardiePi98,ArgamonDaKr99,MPRZ99},
computational biology problems such as identifying splice
sites~\cite{Fickett96,BurgeKarlin98,Haussler98}, and problems in
information extraction~\cite{FreitagMc99}.

Our first approach is within a Markovian framework.  In this case,
classifiers are functions of the observation sequence and their
outcomes represent states; we study two Markov models that are used as
inference procedures and differ in the type of classifiers and the
details of the probabilistic modeling. The critical shortcoming of
this framework is that it attempts to maximize the likelihood of the
state sequence -- not the true performance measure of interest but
only a derivative of it.
The second approach extends a constraint satisfaction formalism
to deal with variables that are associated with costs and shows
how to use this to model the classifier combination problem.  In
this approach general constraints can be incorporated flexibly and
algorithms can be developed that closely address the true global
optimization criterion of interest.
For both approaches we develop efficient combination algorithms that
use general classifiers to yield the inference.

The approaches are studied experimentally in the context of shallow
parsing -- the task of identifying syntactic sequences in
sentences~\cite{Harris57,Abney91,Greffenstette93} -- which has been
found useful in many large-scale language processing applications
including information extraction and text
summarization~\cite{Grishman95,Appeltetal93}.
Working within a concrete task allows us to compare the approaches
experimentally for phrase types such as base Noun Phrases (NPs) and
Subject-Verb phrases (SVs) that differ significantly in their
statistical properties, including length and internal dependencies.
Thus, the robustness of the approaches to deviations from their
assumptions can be evaluated.

Our two main methods, projection-based Markov Models (PMM) and
constraint satisfaction with classifiers (CSCL) are shown to
perform very well on the task of predicting NP and SV phrases,
with CSCL at least as good as any other method tried on these
tasks.
CSCL performs better than PMM on both tasks, more significantly so on
the harder, SV, task.  We attribute it to CSCL's ability to cope
better with the length of the phrase and the long term dependencies.
Our experiments make use of the SNoW classifier \cite{CCRR99,Roth98}
and we provide a way to combine its scores in a probabilistic
framework; we also exhibit the improvements of the standard hidden
Markov model (HMM) when allowing states to depend on a richer
structure of the observation via the use of classifiers.

%------------------------------------------------
\mysection{Identifying Phrase Structure} \label{sec:problem}

The inference problem considered can be formalized as that of
identifying the phrase structure of an input string.
Given an input string $O=<o_1,o_2,\ldots o_n>,$ a {\em phrase} is
a substring of consecutive input symbols $o_{i},o_{i+1},\ldots
o_{j}$. Some external mechanism is assumed to consistently (or
stochastically) annotate substrings as phrases\footnote{We assume
here a single type of phrase, and thus each input symbol is either
in a phrase or outside it. All the methods can be extended to deal
with several kinds of phrases in a string.}. Our goal is to come
up with a mechanism that, given an input string, identifies the
phrases in this string.

The identification mechanism works by using classifiers that
attempt to recognize in the input string local signals which are
indicative to the existence of a phrase. We assume that the
outcome of the classifier at input symbol $o$ can be represented
as a function of the local context of $o$ in the input string,
perhaps with the aid of some external information inferred from
it\footnote{In the case of natural language processing, if the
$o_i$s are words in a sentence, additional information might
include morphological information, part of speech tags, semantic
class information from WordNet, etc.  This information can be
assumed to be encoded into the observed sequence.}.
Classifiers can indicate that an input symbol $o$ is {\bf\em i}nside
or {\bf\em o}utside a phrase (IO modeling) or they can indicate that
an input symbol $o$ {\bf\em o}pens or {\bf\em c}loses a phrase (the OC
modeling) or some combination of the two.
Our work here focuses on OC modeling which has been shown to be more
robust than the IO, especially with fairly long phrases~\cite{MPRZ99}.
In any case, the classifiers' outcomes can be combined to determine
the phrases in the input string. This process, however, needs to
satisfy some constraints for the resulting set of phrases to be
legitimate.  Several types of constraints, such as length, order and
others can be formalized and incorporated into the approaches studied
here.

The goal is thus two fold: to learn classifiers that recognize
the local signals and to combine them in a way that respects the
constraints.  We call the inference algorithm that combines the
classifiers and outputs a coherent phrase structure a {\em
combinator}.
The performance of this process is measured by how accurately it
retrieves the phrase structure of the input string. This is
quantified in terms of {\em recall} - the percentage of phrases that
are correctly identified - and {\em precision} - the percentage of
identified phrases that are indeed correct phrases.

%-----------------------------------------------------
\mysection{Markov Modeling} \label{sec:markov}

HMM is a probabilistic finite state automaton that models the
probabilistic generation of sequential processes.
It consists of a finite set ${\cal S}$ of states, a set ${\cal
O}$ of observations, an initial state distribution $P_1(s)$, a
state-transition distribution $P(s|s')$ ($s, s' \in \mathcal{S}$)
and an observation distribution $P(o|s)$ ($o \in \mathcal{O}$, $s
\in \mathcal{S}$).  A sequence of observations is generated by
first picking an initial state according to $P_1(s)$; this state
produces an observation according to $P(o|s)$ and transits to a
new state according to $P(s|s')$.  This state produces the next
observation, and the process goes on until it reaches a
designated final state~\cite{Rabiner89}.

In a supervised learning task, an observation sequence
$O=<o_1,o_2,\ldots o_n>$ is supervised by a corresponding state
sequence $S=<s_1,s_2,\ldots s_n>$. This allows one to estimate the HMM
parameters and then, given a new observation sequence, to identify the
most likely corresponding state sequence.
The supervision can also be supplied (see Sec.~\ref{sec:problem})
using local signals from which the state sequence can be recovered.
Constraints can be incorporated into the HMM by constraining the state
transition probability distribution $P(s|s')$.  For example, set
$P(s|s') = 0$ for all $s, s'$ such that the transition from $s'$ to
$s$ is not allowed.

%----------------------------------
\mysubsection{A Hidden Markov Model Combinator} \label{sec:hmm}

To recover the most likely state sequence in HMM, we wish to
estimate all the required probability distributions.  As in
Sec.~\ref{sec:problem} we assume to have local signals that
indicate the state. That is, we are given classifiers with states
as their outcomes.  Formally, we assume that $P_t(s|o)$ is given
where $t$ is the time step in the sequence.
In order to use this information in the HMM framework, we compute
$ P_t(o|s) = P_t(s|o)P_t(o)/P_t(s).$
That is, instead of observing
the conditional probability $P_t(o|s)$ directly from training
data, we compute it from the classifiers' output.
Notice that in HMM, the assumption is that the probability
distributions are stationary.  We can assume that for $P_t(s|o)$ which
we obtain from the classifier but need not assume it for the other
distributions, $P_t(o)$ and $P_t(s)$.
$P_t(s)$ can be calculated by $P_t(s) = \sum_{s' \in
\mathcal{S}}P(s|s')P_{t-1}(s')$
where $P_1(s)$ and $P(s|s')$ are
the two required distributions for the HMM.
We still need $P_t(o)$ which is harder to approximate but, for each
$t$, can be treated as a constant $\eta_t$ because the goal is to
find the most likely sequence of states for the given observations,
which are the same for all compared sequences.

With this scheme, we can still combine the classifiers'
predictions by finding the most likely sequence for an
observation sequence using dynamic programming. To do so, we
incorporate the classifiers' opinions in its recursive step by
computing $P(o_t|s)$ as above:
$$\delta_t(s)
= \max_{s' \in \mathcal{S}}\delta_{t-1}(s')P(s|s')P(o_t|s)
 =  \max_{s' \in \mathcal{S}}\delta_{t-1}(s')P(s|s')P(s|o_t)\eta_t/P_t(s).$$
This is derived using the HMM assumptions but utilizes the
classifier outputs $P(s|o)$, allowing us to extend the notion of
an observation. In Sec.~\ref{sec:experiments} we estimate
$P(s|o)$ based on a whole observation sequence rather than $o_t$
to significantly improve the performance.

%----------------------------------
\mysubsection{A Projection based Markov Model Combinator}
\label{projection combinator}
In HMMs, observations are allowed to depend only on the current
state and long term dependencies are not modeled.
Equivalently, the constraints structure is restricted by having a
stationary probability distribution of a state given the previous
one.  We attempt to relax this by allowing the distribution of a
state to depend, in addition to the previous state, on the
observation.
Formally, we now make the following independence assumption:
$P(s_t|s_{t-1}, s_{t-2},\ldots, s_1, o_t, o_{t-1}, \ldots, o_1) =
P(s_t|s_{t-1}, o_t).$
Thus, given an observation sequence $O$ we can find the most
likely state sequence $S$ given $O$ by maximizing
$$P(S|O) = \prod_{t =
2}^{n}[P(s_t|s_1,\ldots,s_{t-1},o)]P_1(s_1|o) = \prod_{t =
2}^{n}[P(s_t|s_{t-1},o_{t})]P_1(s_1|o_1).$$
Hence, this model generalizes the standard HMM by combining the
state-transition probability and the observation probability into
one function.
The most likely state sequence can still be recovered using the
dynamic programming (Viterbi) algorithm if we modify the recursive
step: $\delta_{t}(s) = \max_{s' \in
\mathcal{S}}\delta_{t-1}(s')P(s|s',o_t)$.
In this model, the classifiers' decisions are incorporated in the
terms $P(s|s',o)$ and $P_1(s|o)$. To learn these classifiers we
follow the projection approach~\cite{Valiant98colt} and separate
$P(s|s',o)$ to many functions $P_{s'}(s|o)$ according to the
previous states $s'$. Hence as many as $|\mathcal{S}|$
classifiers, projected on the previous states, are separately
trained. (Therefore the name ``Projection based Markov model
(PMM)''.) Since these are simpler classifiers we hope that the
performance will improve.
As before, the question of what constitutes an observation is an
issue.  Sec.~\ref{sec:experiments} exhibits the contribution of
estimating $P_{s'}(s|o)$ using a wider window in the observation
sequence.

%-----------------------------------------------------
\mysubsection{Related Work} \label{sec:OtherHMMs}

Several attempts to combine classifiers, mostly neural networks, into
HMMs have been made in speech recognition works in the last
decade~\cite{MorganBo95}.  A recent work~\cite{MFP00} is similar to
our PMM but is using maximum entropy classifiers. In both cases, the
attempt to combine classifiers with Markov models is motivated by
an attempt to improve the existing Markov models; the belief is that
this would yield better generalization than the pure observation
probability estimation from the training data.
Our motivation is different.  The starting point is the existence of
general classifiers that provide some local information on the input
sequence along with constraints on their outcomes; our goal is to use
the classifiers to infer the phrase structure of the sequence in a
way that satisfies the constraints.  Using Markov models is only one
possibility and, as mentioned earlier, not one the optimizes the real
performance measure of interest.
Technically, another novelty worth mentioning is that we use a wider
range of observations instead of a single observation to predict a
state.  This certainly violates the assumption underlying HMMs but
improves the performance.

%-----------------------------------------------------
\mysection{Constraints Satisfaction with Classifiers}
\label{sec:csp}

This section describes a different model that is
based on an extension of the Boolean constraint satisfaction (CSP)
formalism~\cite{Mackworth92} to handle variables that are the
outcome of classifiers.
As before, we assume an observed string $O=<o_1,o_2,\ldots o_n>$
and local classifiers that, without loss of generality, take two
distinct values, one indicating {\bf\em o}penning a phrase
and a second indicating {\bf\em c}losing it (OC modeling). The
classifiers provide their output in terms of the probability
$P(o)$ and $P(c)$, given the observation.

We extend the CSP formalism to deal with probabilistic variables (or,
more generally, variables with cost) as follows.
Let $V$ be the set of Boolean variables associated with the
problem, $|V|=n$. The constraints are encoded as clauses and, as
in standard CSP modeling the Boolean CSP becomes a CNF
(conjunctive normal form) formula $f$. Our problem, however, is
not simply to find an assignment $\tau:V \rightarrow \{0,1\}$ that
satisfies $f$ but rather the following optimization problem.
We associate a cost function $c: V \rightarrow \cal{R}$
with each variable, and try to find a solution $\tau$ of $f$ of
minimum cost, $c(\tau) = \sum_{i=1}^{n}\tau(v_i)c(v_i).$

One efficient way to use this general scheme is by encoding
phrases as variables.  Let $E$ be the set of all possible phrases.
Then, all the non-overlapping constraints can be encoded in:
$\bigwedge_{e_i {\mbox ~overlaps~} e_j} (\neg e_i \vee \neg e_j).$
This yields a quadratic number of variables, and the constraints are
binary, encoding the restriction that phrases do not overlap.  A
satisfying assignment for the resulting $2$-CNF formula can therefore
be computed in polynomial time, but the corresponding optimization
problem is still NP-hard~\cite{GusfieldPi92}.
For the specific case of phrase structure, however, we can find
the optimal solution in linear time.
The solution to the optimization problem corresponds to a shortest
path in a directed acyclic graph constructed on the observations
symbols, with legitimate phrases (the variables of the CSP) as its
edges and their cost as the edges' weights.  The construction of
the graph takes quadratic time and corresponds to constructing the
$2$-CNF formula above.
It is not hard to see (details omitted) that each path in this graph
corresponds to a satisfying assignment and the shortest path
corresponds to the optimal solution. The time complexity of this
algorithm is linear in the size of the graph.  The main difficulty
here is to determine the cost $c$ as a function of the confidence
given by the classifiers.  Our experiments revealed, though, that the
algorithm is robust to reasonable modifications in the cost function.
A natural cost function is to use the classifiers probabilities $P(o)$
and $P(c)$ and define, for a phrase $e=(o,c)$, $c(e) = 1-P(o)P(c).$
The interpretation is that the error in selecting $e$ is the error
in selecting either $o$ or $c$, and allowing those to
overlap\footnote{It is also possible to account for the
classifiers' suggestions inside each phrase; details omitted.}.
The constant in $1-P(o)P(c)$ biases the minimization to prefers
selecting a few phrases, so instead we minimize $-P(o)P(c)$.

%------------------------------------------------------------
\mysection{Shallow Parsing} \label{sec:shallow}

We use shallow parsing tasks in order to evaluate our approaches.
Shallow parsing involves the identification of phrases or of
words that participate in a syntactic relationship.  The
observation that shallow syntactic information can be extracted
using local information -- by examining the pattern itself, its
nearby context and the local part-of-speech information -- has
motivated the use of learning methods to recognize these
patterns~\cite{Church88,RamshawMa95,ArgamonDaKr99,CardiePi98}.
In this work we study the identification of two types of phrases,
base Noun Phrases (NP) and Subject Verb (SV) patterns. We chose
these since they differ significantly in their structural and
statistical properties and this allows us to study the robustness
of our methods to several assumptions.
As in previous work on this problem, this evaluation is concerned
with identifying one layer NP and SV phrases, with no embedded
phrases.  We use the OC modeling and learn two classifiers; one
predicting whether there should be an {\bf\em o}pen in location
$t$ or not, and the other whether there should be a {\bf\em
c}lose in location $t$ or not.  For technical reasons the cases
{\bf\em $\neg$o} and {\bf\em $\neg$c} are separated according to
whether we are inside or outside a phrase. Consequently, each
classifier may output three possible outcomes {\bf O}, {\bf nOi},
{\bf nOo} (open, not open inside, not open outside) and {\bf C},
{\bf nCi}, {\bf nCo}, resp.  The state-transition diagram in
figure~\ref{fig:state diagram} captures the order constraints.
Our modeling of the problem is a modification of our earlier work
on this topic that has been found to be quite successful compared
to other learning methods attempted on this problem~\cite{MPRZ99}.
\begin{figure}[!htb]
%\vspace{-0.15 in}
\begin{center}
\epsfig{file=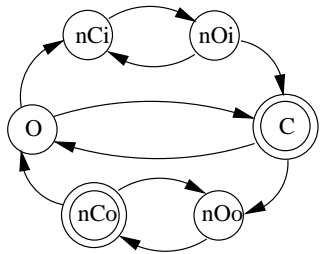} \caption{State-transition diagram for
the phrase recognition problem. \vspace{-0.1in} } \label{fig:state
diagram}
\end{center}
\vspace{-0.15 in}
\end{figure}
%---------------------------------------------

\vspace{-0.1in} \mysubsection{Classification} \label{sec:class}

The classifier we use to learn the states as a function of the
observation is SNoW~\cite{Roth98,CCRR99}, a multi-class classifier
that is specifically tailored for large scale learning tasks. The
SNoW learning architecture learns a sparse network of linear
functions, in which the targets (states, in this case) are
represented as linear functions over a common features space. SNoW
has already been used successfully for a variety of tasks in
natural language and visual
processing~\cite{GoldingRo99,RothYaAh00cvpr}.
Typically, SNoW is used as a classifier, and predicts using a
winner-take-all mechanism over the activation value of the target
classes.  The activation value is computed using a sigmoid
function over the linear sum.  In the current study we normalize
the activation levels of all targets to sum to $1$ and output the
outcomes for all targets (states). We verified experimentally on
the training data that the output for each state is indeed a
distribution function and can be used in further processing as
$P(s|o)$ (details omitted).

%-------------------------------------------------------------
\mysection{Experiments} \label{sec:experiments}

We experimented both with NPs and SVs and we show results for two
different representations of the observations (that is, different
feature sets for the classifiers) - part of speech (POS)
information only and POS with additional lexical information
(words). The result of interest is
$F_{\beta} = (\beta^{2} + 1) \cdot \mbox{Recall} \cdot
\mbox{Precision}/(\beta^{2} \cdot \mbox{Precision} +
\mbox{Recall})$ (here $\beta=1$).
The data sets used are the standard data sets for this
problem~\cite{RamshawMa95,ArgamonDaKr99,MPRZ99} taken from the Wall
Street Journal corpus in the Penn Treebank~\cite{wsj-corpus}.  For NP,
the training and test corpus was prepared from sections 15 to 18 and
section 20, respectively; the SV phrase corpus was prepared from
sections 1 to 9 for training and section 0 for testing.

For each model we study three different classifiers.  The {\em
simple} classifier corresponds to the standard HMM in which
$P(o|s)$ is estimated directly from the data. When the
observations are in terms of lexical items, the data is too sparse
to yield robust estimates and these entries were left empty.
The NB (naive Bayes) and SNoW classifiers use the same feature set,
conjunctions of size $3$ of POS tags (POS and words, resp.) in a
window of size 6.
%------------------------------
\begin{table}[!htb]
\vspace{-0.15 in} \caption{Results ($F_{\beta=1}$) of different
methods on NP and SV recognition} \label{table:results SV/NP}
\begin{center}
{\small
\begin{tabular}{|l|l||c|c||c|c|}
\hline
\multicolumn{2}{|c||}{Method} & \multicolumn{2}{|c||}{NP} & \multicolumn{2}{|c|}{SV} \\
\hline
Model & Classifier & POS tags only & POS tags+words & POS tags only & POS tags+words \\
\hline\hline
 & SNoW & 90.64 & 92.89 & 64.15 & 77.54 \\
HMM & NB & 90.50 & 92.26 & 75.40 & 78.43 \\
 & Simple & 87.83 & & 64.85 & \\
\hline
 & SNoW & 90.61 & 92.98 & 74.98 & 86.07 \\
 PMM & NB & 90.22 & 91.98 & 74.80 & 84.80 \\
 & Simple & 61.44 & & 40.18 & \\
\hline
 & SNoW & 90.87 & 92.88 & 85.36 & 90.09 \\
CSCL & NB & 90.49 & 91.95 & 80.63 & 88.28 \\
 & Simple & 54.42 & & 59.27 & \\
\hline
\end{tabular}
} \vspace{-0.15 in}
\end{center}
\end{table}

The first important observation is that the SV identification
task is significantly more difficult than that the NP task. This
is consistent across all models and feature sets.
When comparing between different models and feature sets, it is clear
that the simple HMM formalism is not competitive with the other two
models.
What is interesting here is the very significant sensitivity to the
feature base of the classifiers used, despite the violation
of the probabilistic assumptions. For the easier NP task, the HMM
model is competitive with the others when the classifiers used
are NB or SNoW.
In particular, the fact that the significant improvements both
probabilistic methods achieve when their input is given by SNoW
confirms the claim that the output of SNoW can be used reliably as
a probabilistic classifier.

PMM and CSCL perform very well on predicting NP and SV phrases
with CSCL at least as good as any other methods tried on these
tasks.
Both for NPs and SVs, CSCL performs better than the others, more
significantly on the harder, SV, task. We attribute it to CSCL's
ability to cope better with the length of the phrase and the long
term dependencies.
%--------------------------------------------------
\mysection{Conclusion} \label{sec:conc}

We have addressed the problem of combining the outcomes of several
different classifiers in a way that provides a coherent inference
that satisfies some constraints. This can be viewed as a concrete
instantiation of the Learning to Reason framework~\cite{KhardonRo97l}.
The focus here is on an important subproblem, the identification
of phrase structure.
We presented two approachs: a probabilistic framework that extends
HMMs in two ways and an approach that is based on an extension of
the CSP formalism.  In both cases we developed efficient
combination algorithms and studied them empirically.
It seems that the CSP formalisms can support the desired
performance measure as well as complex constraints and
dependencies more flexibly than the Markovian approach.  This
is supported by the experimental results that show that CSCL
yields better results, in particular, for the more complex case of
SV phrases.
As a side effect, this work exhibits the use of general classifiers
within a probabilistic framework.
Future work includes extensions to deal with more general
constraints by exploiting more general probabilistic structures
and generalizing the CSP approach.
%--------------------------------------------------

\mysubsubsection{Acknowledgments}

This research is supported by NSF grants IIS-9801638 and
IIS-9984168.
\vspace{-0.1 in}
\small
\bibliographystyle{latex8}
\bibliography{phrases}
\end{document}